\documentclass[10pt,twocolumn,letterpaper]{article}

\usepackage[pagenumbers]{cvpr} 

\usepackage{graphicx}
\usepackage{amsmath}
\usepackage{amssymb}
\usepackage{booktabs}

\usepackage[pagebackref,breaklinks,colorlinks]{hyperref}

\usepackage[capitalize]{cleveref}
\crefname{section}{Sec.}{Secs.}
\Crefname{section}{Section}{Sections}
\Crefname{table}{Table}{Tables}
\crefname{table}{Tab.}{Tabs.}

\def\cvprPaperID{3264} 
\def\confName{CVPR}
\def\confYear{2022}

\begin{document}

\title{Mapping DNN Embedding Manifolds for \\ Network Generalization Prediction}

\author{Molly O'Brien\\
Johns Hopkins University\\
3400 N Charles St, Baltimore, MD\\
{\tt\small mobrie38@jhu.edu}
\and
Julia Bukowski\\
Villanova University\\
800 Lancaster Ave, Villanova, PA\\
{\tt\small julia.bukowski@villanova.edu}
\and
Mathias Unberath\\
Johns Hopkins University\\
3400 N Charles St, Baltimore, MD\\
{\tt\small unberath@jhu.edu}
\and
Aria Pezeshk*\\
Center for Devices and Radiological Health, U.S. FDA\\
Silver Spring, MD\\
{\tt\small Aria.Pezeshk@fda.hhs.gov}
\and
Greg Hager*\\
Dept. of Computer Science, Johns Hopkins University\\
3400 N Charles St, Baltimore, MD\\
{\tt\small hager@jhu.edu}
}

\maketitle

\begin{abstract}
Understanding Deep Neural Network (DNN) performance in changing conditions is essential for deploying DNNs in safety critical applications with unconstrained environments, e.g., perception for self-driving vehicles or medical image analysis.
Recently, the task of Network Generalization Prediction (NGP) has been proposed to predict how a DNN will generalize in a new operating domain. Previous NGP approaches have relied on labeled metadata and known distributions for the new operating domains. In this study, we propose the first NGP approach that predicts DNN performance based solely on how unlabeled images from an external operating domain map in the DNN embedding space.
We demonstrate this technique for pedestrian, melanoma, and animal classification tasks and show state of the art NGP in 13 of 15 NGP tasks without requiring domain knowledge. Additionally, we show that our NGP embedding maps can be used to identify misclassified images when the DNN performance is poor.

\end{abstract}



\section{Introduction}

It is well known that Deep Neural Networks (DNNs) are black box systems that achieve state of the art performance in essentially every perception task proposed in the last decade. DNNs are composed of tens to hundreds of layers with millions of learnable weights, and they excel at tasks such as image classification, object detection, and semantic segmentation.  It is also well documented that DNN performance often degrades when DNNs are deployed in operating domains that are different from the training and testing domains \cite{koh2020wilds}. Because of this performance degradation, even as DNN performance continues to improve and approach human performance in many benchmark datasets, it is challenging to deploy DNNs in commercial products that perform safety critical tasks in unconstrained environments.
In order for DNNs to reach their full potential for commercial use, we need techniques that can predict how a DNN will perform in a new operating domain before it causes automated, harmful failures \cite{obrien2020dependable}.

While DNNs are different from traditional software in that the learned weights cannot be read and interpreted, there is still structure in the mappings that DNNs learn. Feed-forward DNNs perform a high-dimensional, non-linear projection of input data into an embedding space, and the final prediction is a linear projection of the embedding. We are interested in identifying structure in the DNN embedding space as it relates to the DNN performance.
Our primary contribution is a novel NGP method that can accurately predict DNN performance in novel operating domains without requiring prior knowledge or class distributions from the novel operating domains.
Our paper proceeds as follows:
\begin{enumerate}
    \item We fit a decision tree to the DNN embedding space that efficiently maps the manifold of the labeled test data.
    \item We extend a previously proposed NGP algorithm \cite{obrien2021wacv} to predict DNN performance in a novel operating domain based on how the unlabeled operating domain images map into the DNN embedding space.
    \item We evaluate our NGP method on pedestrian, melanoma, and animal classification tasks and demonstrate accurate NGP across different DNN architectures and classification tasks. Additionally, we show that our NGP method can identify misclassified images when the DNN performance is poor.
\end{enumerate}


\section{Background}
DNN performance often degrades when the DNN is deployed in operating domains that are different in some way from the training and testing data \cite{koh2020wilds}.
For instance, in perception for self-driving vehicles, differences in camera characteristics, lighting and weather conditions, and foreground and background objects can impact DNN performance.
In medical image analysis the input data distribution can be impacted by choice of scanner vendors, pre- and post-processing algorithms, dose levels, image compression, and patient and disease distributions.
Ongoing research on improving DNN performance in unconstrained environments focuses on domain generalization, rejecting out-of-distribution (OOD) input images, and predicting how a DNN will perform in a new environment.


\subsection{Domain Generalization}\label{back:domain_generalization}
DNNs trained using domain generalization algorithms aim to perform well in operating domains that differ from the training or testing domains. While many domain generalization algorithms have been proposed in the last decade \cite{arjovsky2019invariant}, none has been shown to consistently out-perform standard Empirical Risk Minimization \cite{gulrajani2020search}.  See \cite{gulrajani2020search} for a review of Domain Generalization techniques.
It has been proposed that underspecification can lead network performance to degrade when deployed in operating domains different from the training  domains \cite{d2020underspecification}.
To facilitate domain generalization research, the WILDS benchmark was released to provide datasets with ``in-the-wild" distribution shifts between the training and test data \cite{koh2020wilds}.

An emerging topic in domain generalization is hidden stratification: the idea that average performance can obscure subpopulations of data where the DNN performs poorly. This can lead to harm if the task is safety critical, e.g., medical image analysis \cite{10.1145/3368555.3384468}. Sohoni et al. propose the framework GEORGE that identifies subgroups of data by clustering examples in the DNN embedding space and training classifiers that demonstrate robust performance across subgroups \cite{NEURIPS2020_e0688d13}.

\subsection{Out-of-Distribution Detection}\label{back:ood}
DNNs are trained in limited environments, e.g., to classify dogs and cats. If a DNN is used in an unconstrained environment, an input sample from outside of that limited training environment, e.g., an image of a bird, standard DNNs will provide an incorrect answer because neither dog nor cat can be correct. Automatically recognizing OOD samples is a broad area of research that is relevant to safely deploying DNNs in unconstrained environments.
Many prior works use the DNN embedding, i.e., the output from the penultimate DNN layer, or the softmax scores to detect OOD samples \cite{DBLP:journals/corr/HendrycksG16c, dcbe7abf4db64d1b89bf9802585660ed, Mohseni_Pitale_Yadawa_Wang_2020, NEURIPS2020_f5496252, 10.1145/3338501.3357372}.
The baseline in \cite{DBLP:journals/corr/HendrycksG16c} uses the softmax scores to predict whether an image is misclassified in addition to OOD detection.
Previous work has investigated input sample Euclidean or Mahalanobis distance from training data in the embedding space to identify OOD and adversarial examples \cite{lee2018simple, 
feinman2017detecting, ma2018characterizing}. 
Recent work proposed the Multi-level Out-of-distribution Detection (MOOD) framework for computationally efficient OOD \cite{Lin_2021_CVPR}.

\subsection{Network Generalization Prediction}\label{back:ngp}
Recent work has proposed Network Generalization Prediction (NGP) as a task of interest for deploying DNNs in unconstrained environments\cite{obrien2021wacv}. The aim of NGP is to predict how a DNN will perform when it is used in a novel operating domain without requiring labeled test data from that domain. Previous work proposed an interpretable context subspace (CS) that identifies context features, i.e., metadata, or image statistics like brightness, that are informative for NGP \cite{obrien2021wacv}. The previous work can accurately predict DNN performance for changes in context feature distribution,  but does not capture changes that occur when moving from one dataset to another, e.g., changes in camera parameters or changes in the image structure.  A similar task, denoted Detection Performance Modeling, was proposed in \cite{ponn2020identification} where Ponn et al. trained a random forest on image attributes, e.g., pedestrian occlusion, bounding box size, presence of rain, etc., to predict whether a pedestrian would be detected.
All previously proposed NGP algorithms require labeled metadata or image statics to predict DNN performance. We propose the first approach that can predict DNN performance directly from how unlabeled images map in the DNN embedding space.

\begin{figure}[t]
\begin{center}\includegraphics[width=.9\linewidth]{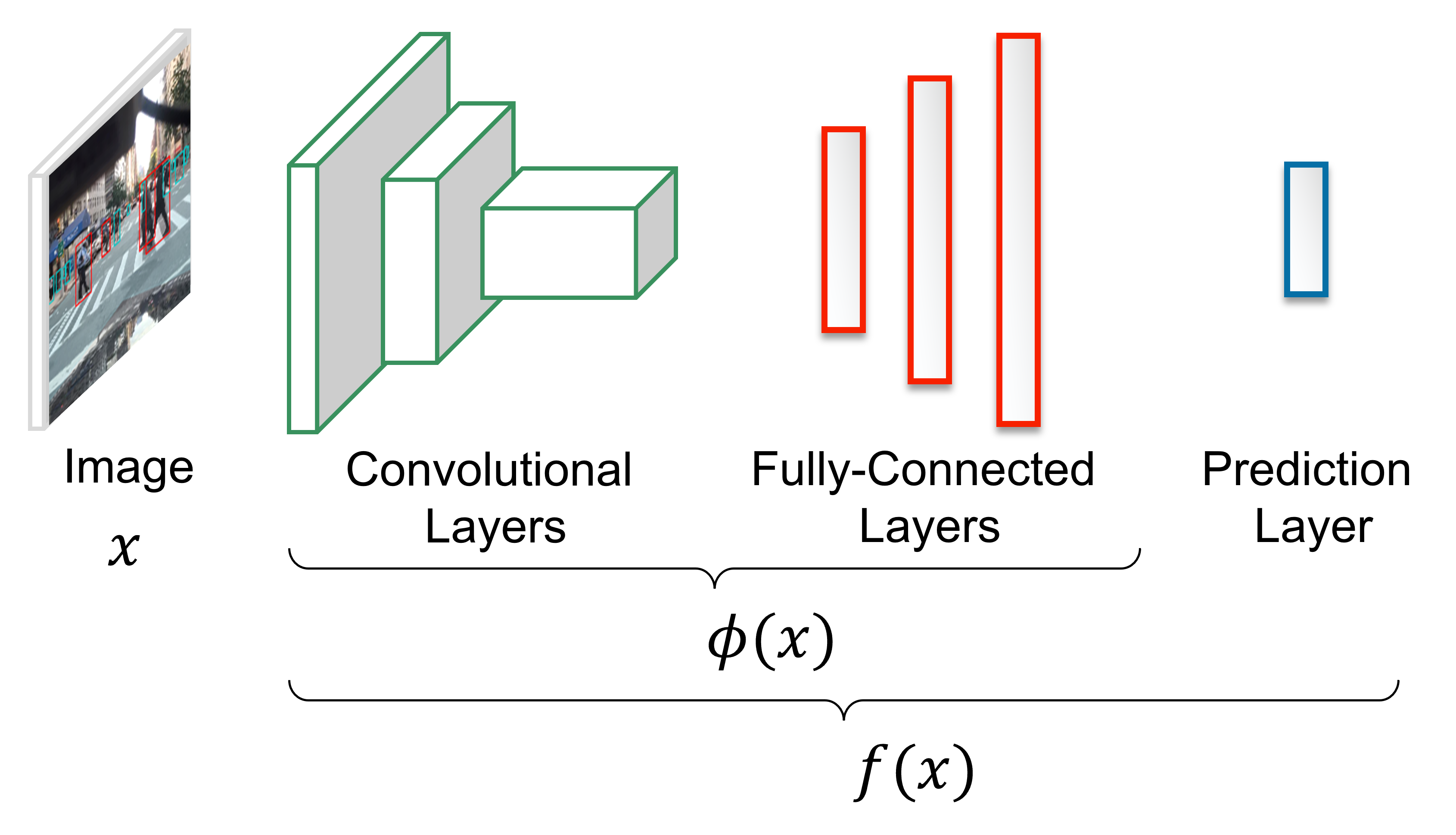}
\end{center}
  \caption{Components of a typical feed-forward Deep Neural Network (DNN): convolutional layers, fully connected layers, and the prediction layer. The prediction layer is also a fully-connected layer that projects the final embedding, $\phi(x)$, into the prediction dimension.}
\label{fig:DNN_notation}
\end{figure}

\section{Methods}
We consider a trained, feed-forward DNN, $f$, where $f(x)$ denotes the DNN prediction, see Figure \ref{fig:DNN_notation}.
The layers of $f$, excluding the final layer, are a feature extractor, denoted $\phi$, that projects the input image $x$ into a $D$ dimensional DNN embedding space (embedding space), $\phi(x) \in R^D$, see Figure \ref{fig:DNN_notation}.
The DNN $f$ is tested with images from an internal test set, i.e., a test set drawn from the same distribution as the training data. The images in the internal test set are denoted $X = \{x_i\}_{i=1}^N$ and are labeled $\textbf{y}=\{y_i\}_{i=1}^N$.
Images from an external operating set $\hat{X} = \{ \hat{x}_i\}_{i=1}^M$ are analogous data from a new distribution.
However, for the external operating set we assume that labels $\hat{\textbf{y}}$ are unknown.

We are interested in finding structure in the embedding space  that provides information about the DNN performance, specifically we aim to link the embedding space to the DNN outcome. The DNN outcome, $o$, can be a function of both the label and the DNN loss. Generally, let $\mathcal{L}(f(x), y)$ denote the loss associated with the DNN prediction $f(x)$ and label $y$.  The outcome is denoted by:
\begin{equation}
    o(\mathcal{L}(f(x), y), y) \triangleq o(f(x),y)
\end{equation}
For simplicity, we use the notation $o(f(x), y)$ to denote the outcome.
Depending on the task, the outcome of interest could be determined by the loss, e.g., success or failure, or by the loss and the label, e.g., for melanoma classification, in addition to modeling success and failure we may want to model misclassifying a malignant image separately from misclassifying a benign image because misclassifying a malignant image is dangerous for the patient. In the experiments in Section \ref{exp} we examine binary classification where there are four possible outcomes, i.e., true positive (TP), false positive (FP), true negative (TN), and false negative (FN).


\begin{figure*}[t]
\begin{center}\includegraphics[width=.9\linewidth]{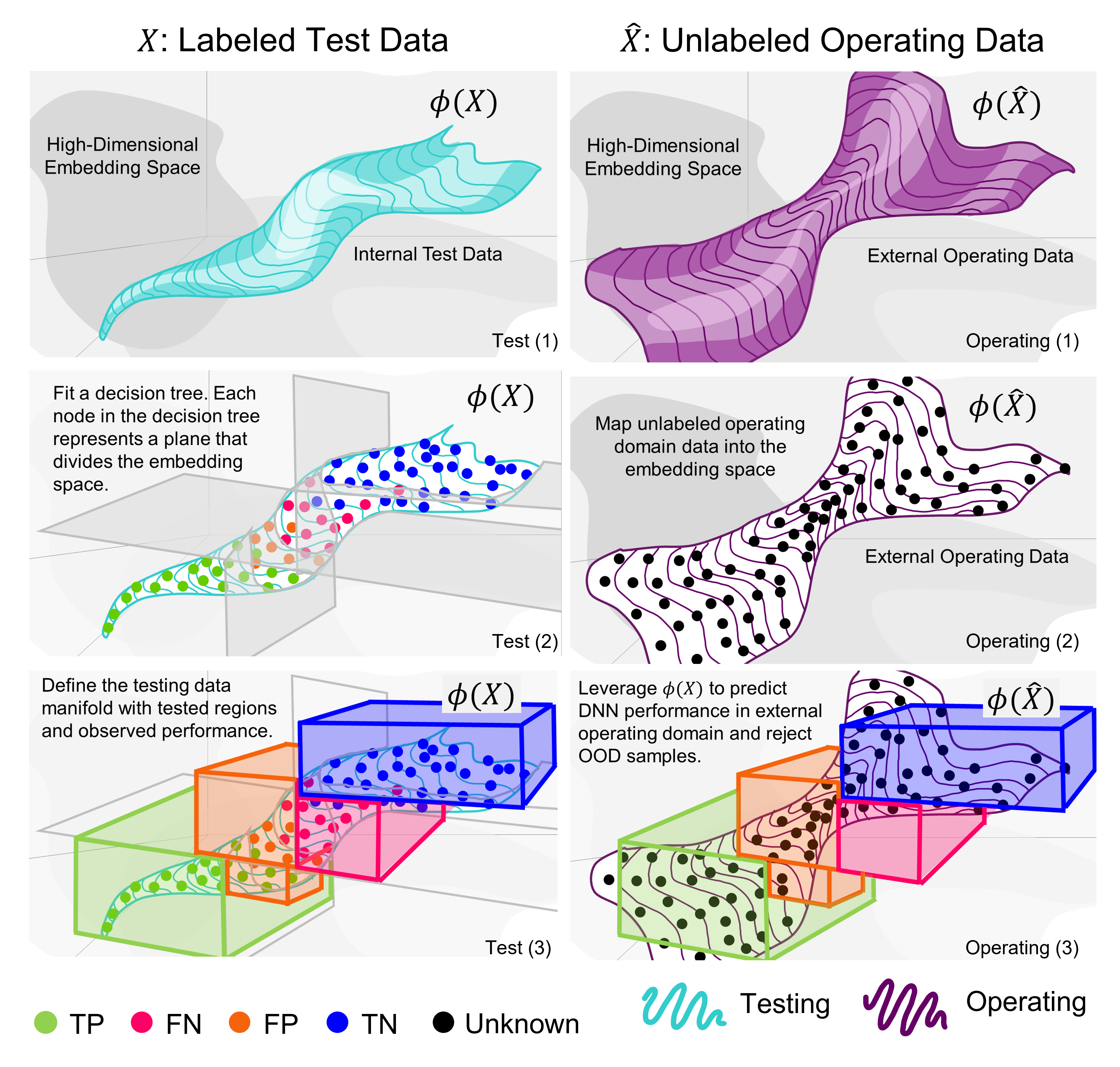}
\end{center}
  \caption{An illustration of the decision tree for mapping DNN embeddings. Test data lie on a manifold in the embedding space. We identify structure in the embedding space as it relates to the DNN outcome. For binary classification the possible outcomes are true positive (TP), false negative (FN), false positive (FP) and true negative (TN). The structure identified using labeled test data can be leveraged to predict the DNN's performance on unlabeled operating data, where the outcome is unknown. Best viewed in color.}
\label{fig:embedding_manifold}
\end{figure*}

\subsection{Decision Tree in Embedding Space}\label{decision_tree}
The internal test set embeddings, $\phi(X)$, lie on some manifold in the high-dimensional embedding space, see Figure \ref{fig:embedding_manifold} Test (1).
Decision trees are able to identify a sparse set of the most informative features given high-dimensional feature data. We fit a decision tree on the $D$-dimensional test set embeddings, $\phi(X)$,  so that the decision tree can predict the observed test outcomes, $o(f(X), \textbf{y})$. We set a maximum depth of the decision tree to prevent the decision tree from overfitting.
The decision tree recursively selects the dimension of the embedding feature that maximizes the information gain about the DNN outcome. After the decision tree is fit, each node in the tree corresponds to a hyper-plane in the embedding space, see Figure \ref{fig:embedding_manifold} Test (2). Each leaf node in the tree corresponds to a contiguous region in embedding space identified using a sparse subset of the embedding dimensions that give the most information (in a greedy sense) about the DNN outcome. We refer to the fitted decision tree as our embedding map: it maps regions in the embedding space to observed DNN outcomes.

\subsection{Approximating Internal Test Set Manifold}\label{manifold_approx}
The embedding map found in Section \ref{decision_tree} contains $L$ leaf nodes that define contiguous regions in the embedding space, where leaf $l$ is identified using a sparse subset $\textbf{d}^l << D$ of the embedding space dimensions. Note that the number of dimensions in $\textbf{d}^l$ is less than or equal to the maximum depth of the decision tree. Using the embedding map, we can partition the internal test samples $X = \{x_i\}_{i=1}^N$ by the leaf to which each sample maps, i.e.,
\begin{equation}
   X = \cup_{l=1}^L X^l , \quad \quad X^i \cap X^j = \emptyset \quad \forall i \neq j
\end{equation}
where $X^l$ is the set of $N^l$ test samples that map to leaf $l$.
\begin{equation}
    X^l = \{x_i^l\}_{i=1}^{N^l}
\end{equation}
We want link the embedding space to the outcomes observed in testing and identify the tested regions of the embedding space. Given the contiguous region in the embedding space defined by leaf $l$ and the test samples $X^l$ that map to leaf $l$, we can define the tested region in the embedding space as a convex hull, $H^l$, around  $\phi(X^l)$. Let $\phi(X)[d]$ indicate the $d^{th}$ dimension of the embedding feature. Then $H^l$ is given by:
\begin{equation}
    H^l = [min(\phi(X^l)[d]), max(\phi(X^l)[d])] \quad \forall d \in \textbf{d}^l
\end{equation}
See Figure \ref{fig:embedding_manifold} Test (3) for an illustration of convex hulls around the test samples in the embedding space.

The internal test set $X$ has associated labels $\textbf{y}$. Let $\textbf{y} = \cup_{l=1}^L \textbf{y}^l$ be the test set labels partitioned by the leaf to which each sample maps. $\textbf{y}^l = \{y_i^l \}_{i=1}^{N^l} $ are
the labels for the test samples that map to leaf $l$.
Let $\mathbb{I}(\textbf{a},\textbf{b})$ be an indicator function that is equal to $1$ if $\textbf{a} = \textbf{b}$ and $0$ otherwise.
Assuming each test sample is equally likely, the probability of outcome $a$ in leaf $l$ can be computed as:
\begin{equation}
    p(a| l) = \frac{1}{N^l} \sum_{i=1}^{N^l}{\mathbb{I}(o(f(x_i^l), y_i^l), a )}
\end{equation}
The boxes in Figure \ref{fig:embedding_manifold} Test (3) are colored to match the most likely test outcome in the leaf region to illustrate linking a region of embedding space to the DNN outcomes observed in testing.

\subsection{Inference on External Operating Data}
We leverage the embedding map on unlabeled, external operating data, see Figure \ref{fig:embedding_manifold} Operating (1). The external operating samples can be mapped into the DNN embedding space as $\phi(\hat{X})$, see Figure \ref{fig:embedding_manifold} Operating (2).
For the operating samples that map to leaf $l$, the sample is deemed ``inside" the testing domain if it lies inside the convex hull of the test samples observed at that leaf node, $H^l$:
\begin{equation}
    \phi(\hat{x})[d] \in [min(\phi(X^l)[d]), max(\phi(X^l)[d])] \quad \forall d \in \textbf{d}^l
\end{equation}
The external operating samples that do not map inside the tested regions are assigned to leaf $L+1$ where the outcome is unknown. The external operating samples can then be partitioned to the $L+1$ leaf nodes:
\begin{equation}
   \hat{X} =  \cup_{l=1}^{L+1} \hat{X}^l, \quad \quad \hat{X}^i \cap \hat{X}^j = \emptyset \quad \forall i \neq j
\end{equation}
where $\hat{X}^l$ is the set of $M^l$ external operating samples that map to leaf $l$.
\begin{equation}
    \hat{X}^l = \{\hat{x}_i^l\}_{i=1}^{M^l}
\end{equation}

\subsection{Network Generalization Prediction}
The goal of Network Generalization Prediction is to predict DNN performance for an unlabeled, external operating set from which labeled test data is not available. The probability that a sample in the external operating set maps to leaf $l$ can be approximated by the fraction of the operating samples that map to leaf $l$:
\begin{equation}
p(l) = \frac{M^l}{M}
\end{equation}
The probability of encountering outcome $a$ in the operating domain is:
\begin{equation}\label{eq:outcome_prob}
    p(a) = \sum_{l \in L} p(a | l) p(l)
\end{equation}
$p(a)$ can be computed for each outcome $a$ observed in testing (assuming we have discrete outcomes and outcome possibilities).
The probability of an unknown outcome can be computed as the probability that external operating samples map outside the tested regions, i.e., $M^{L+1}/M$.



\begin{figure*}[t]
\begin{center}\includegraphics[width=.9\linewidth]{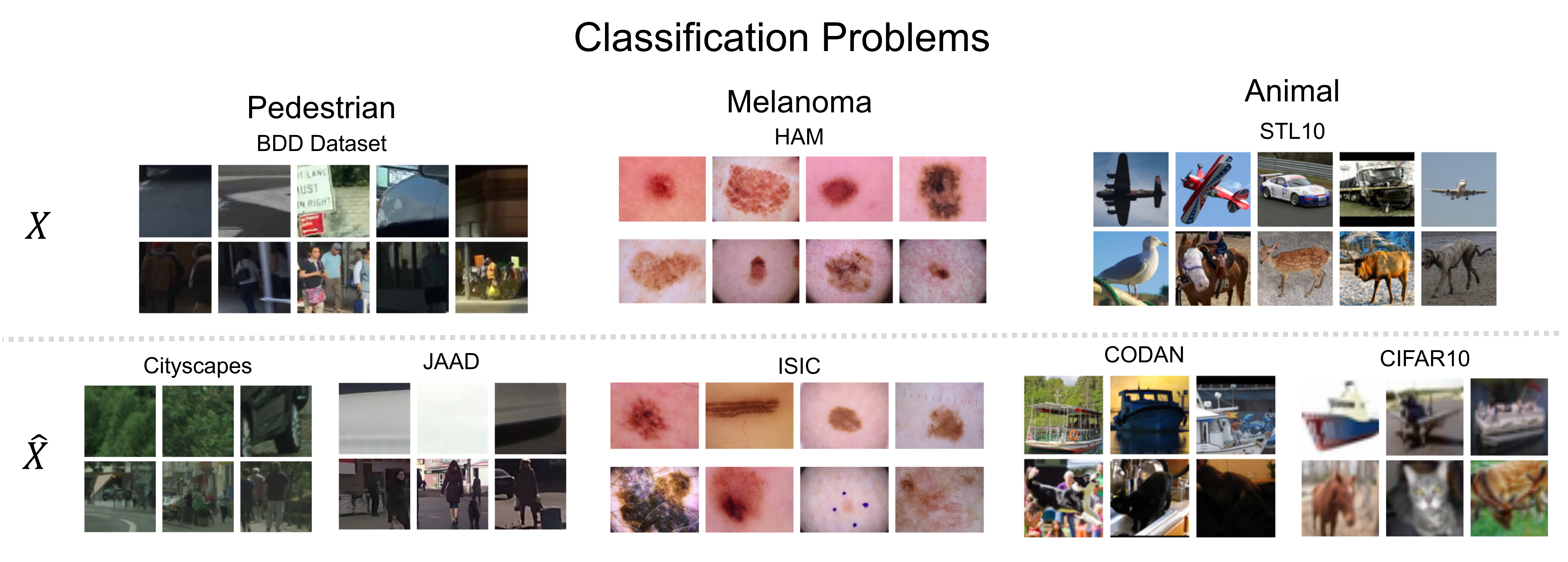}
\end{center}
  \caption{Classification tasks. $X$ indicates the internal dataset that is used to train the DNN classifier and fit the embedding decision tree. $\hat{X}$ indicates the unlabeled, external operating dataset. For each dataset, the top row shows a random sampling of negative examples, and the bottom row shows a random sampling of positive examples. }
\label{fig:tasks}
\end{figure*}


\section{Experiments}\label{exp}
We perform Network Generalization Prediction for three classification tasks: pedestrian classification, melanoma classification, and animal classification.
\subsection{Pedestrian Classification}
One of the most safety-critical tasks for autonomous perception systems in self-driving vehicles is to detect and avoid pedestrians. We consider pedestrians from three driving perception datasets: Berkeley Deep Drive 100k (BDD) \cite{yu2018bdd100k}, Cityscapes  \cite{Cordts2016Cityscapes}, and Joint Attention in Autonomous Driving (JAAD) \cite{rasouli2017ICCVW}. From the images in these datasets, for positive examples, we cropped square patches containing pedestrians, with area of greater than 300 pixels.
For negative examples, we cropped random, square image patches of 100 $\times$ 100 pixels. The pedestrian image patches were resized to 100 $\times$ 100 pixels. We use BDD as the internal dataset versus Cityscapes and JAAD that are used as external datasets. The BDD dataset was recorded in different settings across the US in varying weather conditions, day and night times. Cityscapes was recorded in 50 cities in Germany during the day in fair weather conditions. JAAD was recorded in North America and Europe in mainly daytime settings with clear weather. Between the internal and external datasets, we expect changes in image statistics like brightness and saturation as well as some structural changes in the size and location of pedestrians in the image due to changes in how roads and sidewalks are laid out in different cities and countries.

\subsection{Melanoma Classification}
Melanoma is a deadly and fast-moving skin cancer. Early detection of melanoma is key for effective treatment, so there is significant interest in automated techniques like smart phone apps that can screen for melanoma. After a DNN is deployed, it will be exposed to images of skin lesions from new operating domains that are subject to many changes compared to the original training and test data such as changes in lighting conditions, camera settings, and patient populations. Therefore, accurate NGP is needed to understand how a DNN will perform in a new setting and for different subpopulations. We address the task of classifying an image of a skin lesion as melanoma (the positive class), or benign (the negative class). We use the Human Against Machine 10000 (HAM) dataset \cite{tschandl2018ham10000} for our internal dataset and the SIIM-ISIC Melanoma Classification (ISIC) dataset \cite{rotemberg2021patient} for our external dataset. HAM images are $450\times600$ pixels. We resized ISIC images to be the same size, i.e., $450\times600$ pixels.

\subsection{Animal Classification}
General animal and object classification is a standard computer vision task that could have safety implications in the future, e.g., household autonomous robots will need to differentiate pets from inanimate objects in order to safely navigate around changing environments.
We aim to classify an image as an animal (the positive class) or an object (the negative class), with STL10 \cite{stl10} as the internal dataset. The STL10 images are 96 $\times$ 96 pixels and they include animals: birds, horses, deer, dogs, and cats, and objects: planes, cars, trucks, and boats. For external datasets we use the Common Objects Day and Night (CODaN) \cite{codan} and CIFAR-10 \cite{cifar10} datasets. In our NGP experiments, we only include the external dataset images of classes seen during training.  The CODaN animal classes considered are dogs and cats, and the CODaN object classes considered are cars and boats. The CODaN dataset was compiled from other existing datasets; we exclude images taken from ImageNet because the DNN classifiers we fine-tuned were originally trained on ImageNet. The CODaN dataset includes night and day images that are 256 $\times$ 256 pixels; we resized the images to 96 $\times$ 96 pixels. The CIFAR-10 animal classes considered are birds, horses, dogs, cats, and deer. The CIFAR-10 object classes considered are boats, cars, and trucks. CIFAR-10 images are 32 $\times$ 32 pixels, which we resized to 96 $\times$ 96 pixels.  Note that for animal classification, both external datasets present some significant change in image distribution: the CODaN dataset includes night images that were not present in the internal dataset and the CIFAR-10 dataset has images with $1/3$ the resolution of the internal dataset.

\subsection{Distribution Shifts}
In our experiments, we investigate DNN generalization in the midst of diverse external operating domain image shifts. For pedestrian classification, the BDD dataset was recorded across the US in city, highway and residential scenes while Cityscapes was recorded in German cities and JAAD was recorded in North America and Europe. Between the internal and external pedestrian datasets, we expect changes in image appearance, e.g., brightness and saturation, as well as structural image changes, e.g., changes in the size and location of pedestrians in the image due to changes in road and sidewalk layouts in different countries. For melanoma classification, the ISIC operating data has more variation in image appearance, e.g., image saturation and hue, than the internal dataset. For animal classification, the internal dataset STL10 only includes daytime images while the CODaN dataset includes day and night images. The CIFAR-10 images are 1/3 the resolution of the internal images.

\subsection{Experimental Setup}\label{exp_setup}
For each classification task, we fine-tune three classifiers with different DNN architectures: VGG \cite{vgg}, AlexNet \cite{alexnet}, and DenseNet \cite{densenet}; the pre-trained models are available in the PyTorch library.
Each round of training considers 100 batches of images with a batch size of 8, where the images are sampled with a uniform probability for each class.
The VGG and AlexNet models are trained with 10 rounds of training, a learning rate of $1e-6$ and a weight decay of $1e-3$. The DenseNet models are trained with 4 rounds of training, a learning rate of $1e-4$, and a weight decay of  $1e-3$. VGG and AlexNet have an embedding space of $4,096$ dimensions. DenseNet projects into an embedding space of $W \times H \times 1664$ where $W$ and $H$ depend on the initial image size. Like the full DenseNet architecture, we use a Global Average Pooling (GAP) layer to convert from the 3D embedding to a $1664$ dimensional vector for each image. For each architecture in each task, we fit a decision tree with a maximum depth of 10 to distinguish four outcomes: TP, FP, FN, and TN. We refer to the fitted decision tree as our embedding map.

\subsection{Network Generalization Prediction}
We pass the external dataset through the DNN to obtain the embeddings, $\phi(\hat{X})$. We subsequently map the external embeddings to leaves in the embedding map, $ \hat{X} = \cup_{l=1}^{L+1} \hat{X}^l $. For each outcome, TP, FN, FP, and TN, we compute the probability of observing the outcome in the external dataset according to equation \ref{eq:outcome_prob}. If the external dataset has images that map outside the tested regions, i.e., outside the convex hulls defined as $H^l$ for each leaf, the expected outcome is unknown.
The true results are the classification results when the DNN is evaluated using the external dataset labels.

We present the NGP results in two ways: numerically using an F1 score and visually. To facilitate numerical comparison, we show NGP results with an F1 score \cite{lipton2014thresholding} where we compare the predicted probability of a correct outcome and predicted probability of failure with the true probability of a correct outcome and the true probability of failure.
For our NGP algorithm, it is ambiguous whether the unknown outcomes will be correct or a failure. To address this, we compute the F1 score in two ways: assuming the unknown outcomes are correct classifications, denoted ``Ours" in Table \ref{tab:NGP_results}, and assuming the unknown outcomes are failures, denoted ``Ours+" in Table \ref{tab:NGP_results}. The two assumptions reflect optimistic and conservative failure probability predictions.
The F1 score shows whether the NGP algorithm accurately predicts overall success and failure of the DNN.

\begin{figure}[b]
\begin{center}\includegraphics[width=.9\linewidth]{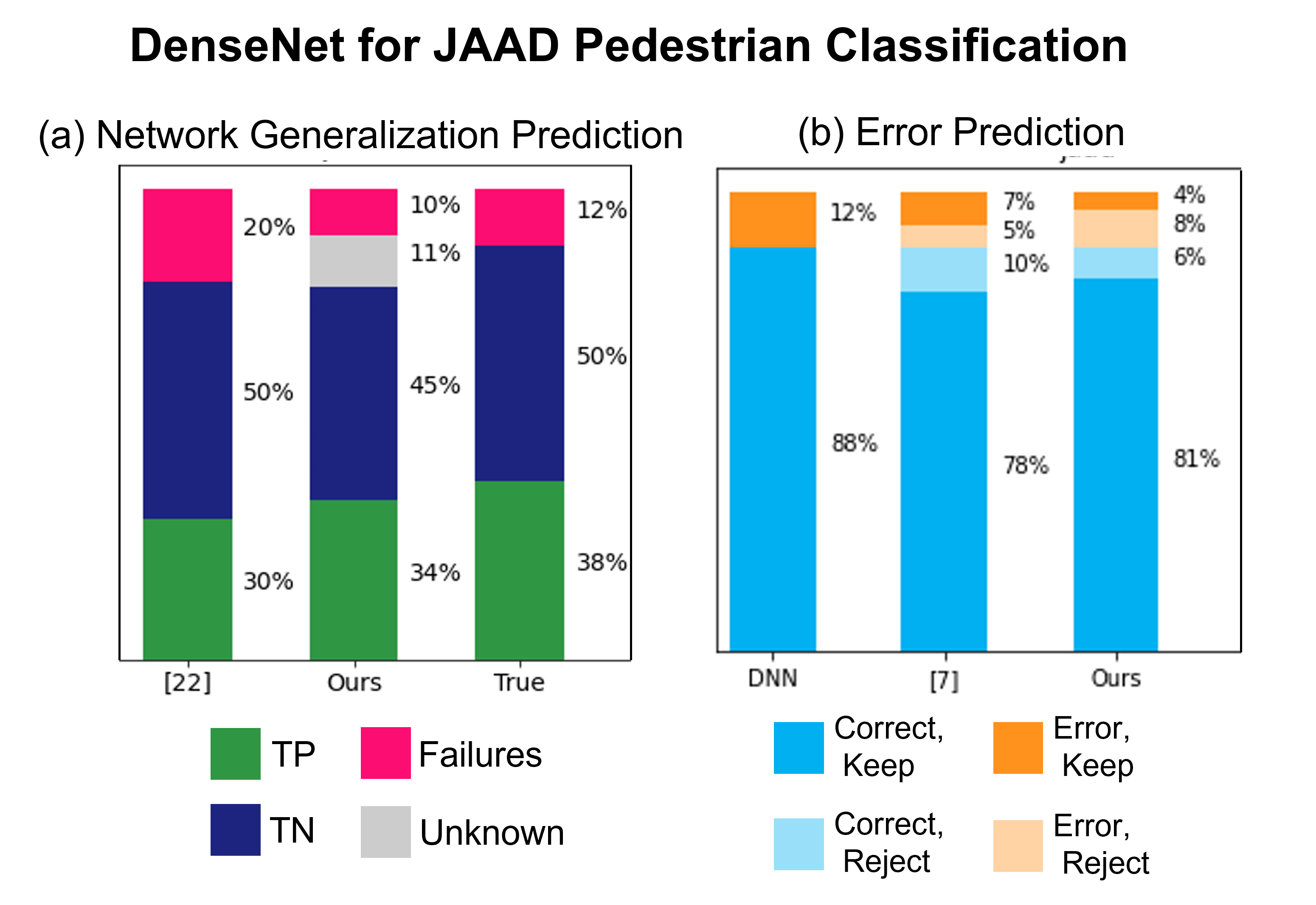}
  \caption{DenseNet JAAD visualized results for (a) Network Generalization Prediction, (b) Error Prediction.}
\label{fig:jaad_densenet}
\end{center}
\end{figure}

\begin{figure*}[t]
\begin{center}\includegraphics[width=.9\linewidth]{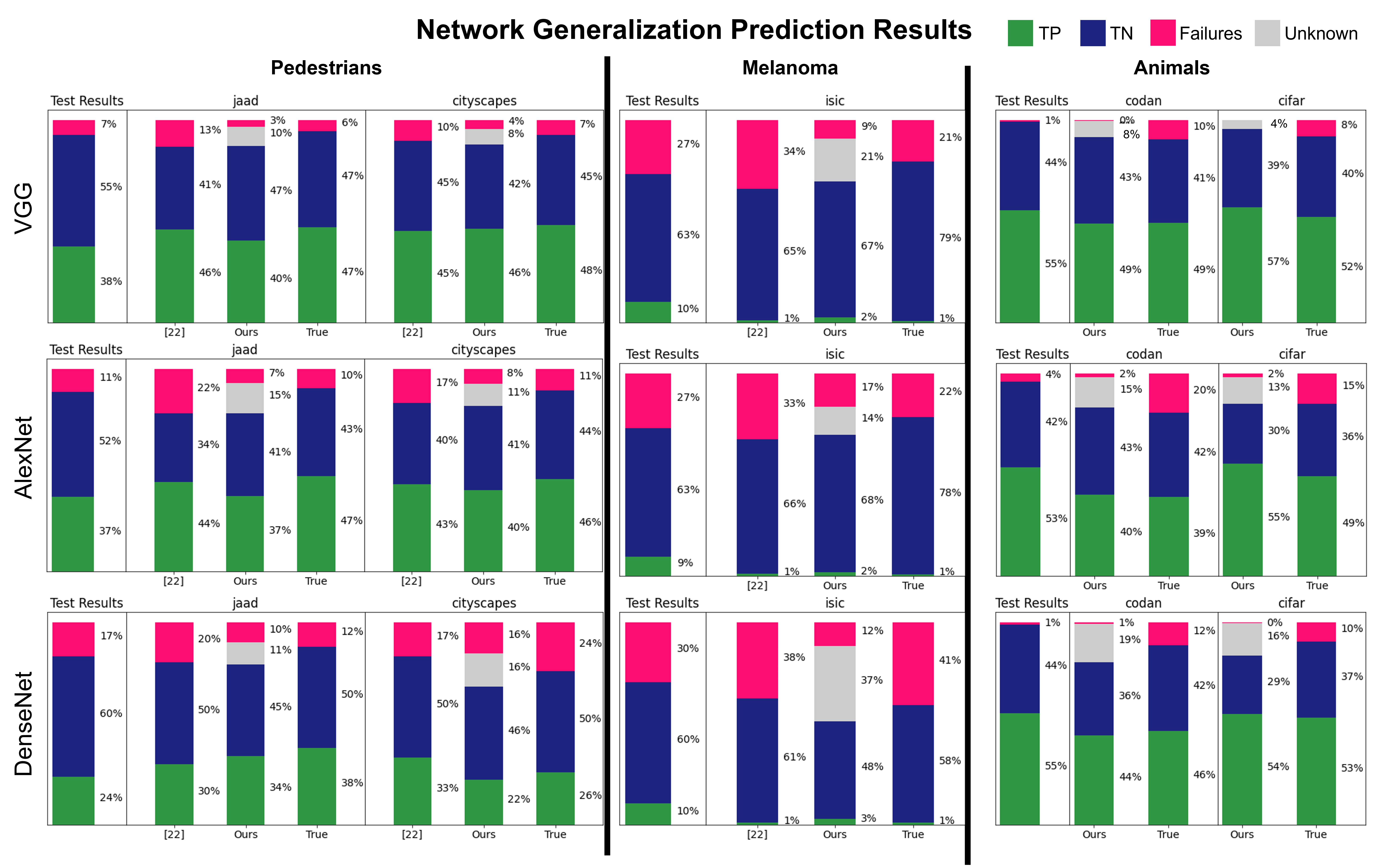}
  \caption{Network Generalization Prediction results for pedestrian classification, melanoma classification, and animal classification. We show results for three DNN architectures: VGG, AlexNet, and DenseNet. }
\label{fig:NGP}
\end{center}
\end{figure*}

Second, we present the full granularity of the NGP results visually in Figure \ref{fig:jaad_densenet} and Figure \ref{fig:NGP}. The stacked bar graphs show the predicted probability of
TP, TN, failure, and an unknown outcome. The goal is to have the predicted results match the `True' results, i.e., the classification results when the DNN is evaluated using the external dataset images and labels. It is common to use the internal test results to predict DNN generalization, so we show the average test set results as a naive baseline for performance prediction, denoted Test Results in Figure \ref{fig:NGP}.

We compare against the NGP approach proposed in \cite{obrien2021wacv}, denoted CS NGP. The CS NGP requires distributions of each class and the distribution of context features to make predictions. For pedestrian classification we use image brightness, scene type, weather, and time of day as available context features. For melanoma classification we use average image hue, saturation, value, patient age, sex, and lesion location as possible context features. Labeled metadata are not available for the animal classification task so we cannot include a comparison to CS NGP.

\begin{table}
\centering
\begin{tabular}{|c c | c c c |}
 \hline
Operating      &   &  & Architecture &   \\
Domain  & NGP & VGG & AlexNet & DenseNet \\ [0.5ex]
 \hline\hline
  & \cite{obrien2021wacv} & 0.984 & 0.964 & \textbf{0.956}  \\

 Cityscapes & Ours & \textbf{0.985} & \textbf{0.984} & 0.947  \\

    & Ours+ & 0.972 & 0.954 & 0.945  \\
 \hline
    & \cite{obrien2021wacv} & 0.958 & 0.927 & 0.954  \\

 JAAD & Ours  & \textbf{0.988} & \textbf{0.986} & \textbf{0.987}  \\

    & Ours+ & 0.959 & 0.927 & 0.948  \\
 \hline
 \hline
    & \cite{obrien2021wacv} & 0.908 & 0.923 & \textbf{0.971}  \\

 ISIC & Ours & \textbf{0.935} & \textbf{0.970} & 0.800  \\

    & Ours+ & \textbf{0.935} & 0.940 & 0.929  \\
 \hline
 \hline

 CODaN & Ours & 0.950 & 0.901 & 0.942  \\

    & Ours+ & \textbf{0.993} & \textbf{0.984} & \textbf{0.951}  \\
 \hline

 CIFAR-10 & Ours & 0.959 & 0.928 & 0.951  \\

    & Ours+ & \textbf{0.982} & \textbf{1.000} & \textbf{0.960}  \\
 \hline
\end{tabular}
\caption{NGP numerical F1 results for pedestrian, melanoma, and animal classification tasks with different architectures.}\label{tab:NGP_results}
\end{table}

\subsection{Network Generalization Prediction Results}

In Table \ref{tab:NGP_results} we show the numerical F1 NGP results. Our proposed NGP approach can robustly predict performance over different DNN architectures, classification tasks, and different external dataset distributions. The proposed NGP algorithm consistently makes more accurate predictions than the CS NGP baseline in \cite{obrien2021wacv} and does not require knowledge about class or context distributions. Our approach is state of the art in 13 of the 15 examples while the CS NGP baseline is  more accurate only when predicting performance for Cityscapes, DenseNet and ISIC, DenseNet.

In Figure \ref{fig:jaad_densenet} we show a sample of the NGP visualized results for the JAAD external dataset with the DenseNet architecture, so that the results can be clearly explained. The NGP visualized results for all operating domains and DNN architectures are shown in Figure \ref{fig:NGP}. In Figure \ref{fig:jaad_densenet} and Figure \ref{fig:NGP} the  pink bars represent the probability of a failure, i.e., FN or FP, and for our algorithm, the gray bar represents the probability of an unknown outcome.

In Table \ref{tab:NGP_results} the predictions marked ``Ours" correspond to predicting that the failures shown in pink are failures and the unknown outcomes are correct. In Table \ref{tab:NGP_results} the predictions marked ``Ours+" correspond to predicting that the failures shown in pink are failures and predicting the unknown outcomes are also  failures. Note that for JAAD, DenseNet the predicted probability of failure is $10\%$,
which is close the true probability of failure of $12\%$,  while the probability of failure or an unknown outcome is $21\%$, which over-predicts failure. Correspondingly ``Ours" is the most accurate prediction for JAAD, DenseNet.  Note that the true probability of failure lies between the predicted probability of failure and the predicted probability of failure plus the probability of an unknown outcome.
Looking at Figure \ref{fig:NGP}, it can be seen that this is true for all tasks and all architectures except VGG for animal classification where we slightly under-predict failure.
Additionally, our DNN performance predictions are consistently more accurate than both baseline methods, CS NGP and the average test results, respectively.


The CS NGP method requires labeled distributions for the probability of the positive and negative classes where our method does not. Note that from the embedding space we are able to capture not only the probability of failure but also the overall class distribution for the external domain whether or not those distributions are the same. In the pedestrian and animal examples, the probability of positive and negative samples is roughly equivalent. On the other hand, in the melanoma example  it is far more likely to have a benign image than an image of melanoma. In all tasks we capture accurate overall class distribution for the external operating domain.

\subsection{Error Prediction}
In the NGP task we predict the probabilities of different outcomes for the entire operating domain. We can also leverage the embedding map to predict whether individual images have been misclassified. For the error prediction task, we denote correct predictions as the positive class and incorrect predictions as the negative class. From our embedding map we predict that operating samples that map to a leaf node with a probability of failure greater than $50\%$ will be misclassified, resulting in $n$ rejected samples. As it is ambiguous whether the unknown outcomes will be correct or a failure, we assume that images that map to an unknown outcome will be correctly classified. To compare against a baseline, we use the ranking in \cite{DBLP:journals/corr/HendrycksG16c} and predict that the $n$ lowest scoring samples will be misclassified. We report error prediction results using the F1 score.

\subsubsection{Error Prediction Results}

We present the error prediction results in Table \ref{app:error_pred_table} for the pedestrian classification, melanoma classification, and animal classification tasks.
Note, neither our proposed error prediction approach nor the baseline outperform keeping all the DNN predictions in many experiments. However, our proposed error prediction approach outperforms both the DNN and the baseline specifically when the overall DNN performance is poor.
In Table \ref{tab:error_pred} we reprint the results from Table \ref{app:error_pred_table} for the operating domains and the DNN architectures where the observed probability of failure throughout the operating domain is greater than or equal to $15\%$.
When the DNN performance is poor, our error prediction approach outperforms the DNN and the baseline in three of five experiments. However, for the animal classification task, there are two examples where the DNN performance is poor and our NGP algorithm does not predict a high probability of failure: CODaN, AlexNet ($20\%$ probability of failure) and CIFAR-10, AlexNet ($15\%$ probability of failure). In these examples we predict an unknown outcome with a probability of $15\%$, and $13\%$, respectively (see Figure \ref{fig:NGP} for the NGP predictions for all tasks and architectures). The external datasets for animal classification represent two of the largest distribution shifts we encounter: CODaN includes day and night images where the internal dataset only includes daytime images. CIFAR-10 images are originally $32 \times 32$ pixels but we resize them to be
$96 \times 96$ to match the size of the internal dataset images; this is a dramatic reduction in image resolution for the DNN.  It is not surprising that most of the failures stem from images that map outside the tested region, but this means that our proposed error prediction method is not as effective for very large domain shifts.

\begin{table}
\centering
\begin{tabular}{|c c | c c c |}
 \hline
Operating      &   &  & Architecture &   \\
Domain  &  E.P. & VGG & AlexNet & DenseNet \\ [0.5ex]
 \hline\hline
  & DNN & \textbf{0.962} & \textbf{0.944} & 0.863 \\
  & \cite{DBLP:journals/corr/HendrycksG16c} & 0.954 & 0.932 & 0.809  \\

 Cityscapes & Ours & 0.949 & 0.920 & \textbf{0.902}  \\


 \hline
   & DNN & \textbf{0.971} & \textbf{0.949} & 0.935 \\
   & \cite{DBLP:journals/corr/HendrycksG16c} & 0.961 & 0.932 & 0.901  \\

 JAAD & Ours & 0.952 & 0.915 & \textbf{0.939}  \\

 \hline
 \hline
   & DNN & 0.885 & 0.879 & 0.741  \\

   & \cite{DBLP:journals/corr/HendrycksG16c} & 0.845 & 0.834 & 0.646 \\

  ISIC & Ours & \textbf{0.905} & \textbf{0.928} & \textbf{0.822}  \\

 \hline
 \hline
  & DNN & \textbf{0.949} & \textbf{0.891} & \textbf{0.938}  \\

   & \cite{DBLP:journals/corr/HendrycksG16c} & 0.946 & 0.880 & 0.936   \\

 CODaN & Ours &  0.945 & 0.867 & 0.923  \\

 \hline
    & DNN & \textbf{0.958} & \textbf{0.918} & \textbf{0.949}  \\
   & \cite{DBLP:journals/corr/HendrycksG16c} & 0.957 & 0.910 & 0.943  \\

 CIFAR-10 & Ours & 0.954 & 0.899 & 0.933  \\

 \hline
\end{tabular}
\caption{\label{app:error_pred_table} Error prediction (E.P.) F1 scores for pedestrian, melanoma, and animal classification tasks.}
\end{table}

\begin{table*}\label{sf:error_pred}
\centering
\begin{tabular}{|c | c | c c c | c | c |}
 \hline
OD     & Cityscapes   &  &  ISIC &  &   CODaN & CIFAR-10\\
Arch.        & DenseNet  & VGG & AlexNet & DenseNet & AlexNet & AlexNet\\
\hline
    DNN      & 0.863  & 0.885 & 0.879 & 0.741 & \textbf{0.891} & \textbf{0.918} \\
    \cite{DBLP:journals/corr/HendrycksG16c} & 0.809 &   0.845 & 0.834 & 0.646 & 0.880 & 0.910 \\
  Ours & \textbf{0.902}  & \textbf{0.905} & \textbf{0.928} & \textbf{0.822} & 0.867 & 0.899 \\
 \hline
\end{tabular}
\caption{\label{tab:error_pred} Error prediction  F1 scores for classification tasks where the probability of failure in the operating domain (OD) and architecture (Arch.) is greater than or equal to $15\%$.}
\end{table*}

\section{Discussion}
We demonstrate that mapping the structure in the DNN embedding space can lead to powerful prediction of DNN performance in external datasets across three high-complexity perception tasks. We consider pedestrian classification and melanoma classification, both of which are tasks where DNNs must perform well in unconstrained environments and where commercial products are available at present. Techniques that can accurately predict DNN performance in new operating domains without requiring labeled data, like our proposed technique, are essential for both the safety and the fairness of DNNs, see Section \ref{disc:impact} for a more thorough discussion of the societal impact of our proposed method.

In addition, we demonstrate that by mapping the embedding space structure we can perform error prediction and improve the DNN performance when the DNN has a high probability of failure. When the DNN performance is poor, i.e., when error prediction is most important,  our proposed error prediction method improves overall performance over the DNN and the baseline. With our embedding map we are leveraging information about the structure of the embedding space. This is fundamentally different than leveraging information from the softmax scores and can be a complementary source of information.

DNN generalization depends on the internal training and testing data, the DNN, and the operating domain. Our proposed method does not require labeled operating data, can predict accurately how a DNN will generalize, and can improve overall performance when the DNN performance is poor. The percent of the operating data where the outcome is unknown can be used to determine whether it is appropriate to deploy the DNN in the novel operating domain  and answer questions like \textit{has the DNN been sufficiently tested?} and \textit{have the right tests been performed?}

Our proposed approach is not restricted to binary classification problems, and is applicable for other feed-forward supervised learning problems, such as, multi-class classification and object detection.
We leverage the structure in the DNN embedding space for NGP and error prediction, but this structure is likely useful for other tasks like OOD rejection. Other directions for future work include more investigation on the decision tree structure. We began our experimentation with a tree depth of 10 arbitrarily and obtained exceedingly good results. The decision trees converged without finding the maximum number of leaves possible, so we did not do extensive experiments with different tree depths.
Future work could investigate different decision tree depths or random forests.



\subsection{Societal Impact}\label{disc:impact}
While we do not address the area of fairness in AI directly, training DNNs that are fair to different subpopulations is essential to safely deploy DNNs in unconstrained environments. There is evidence that both pedestrian detection and melanoma classification can have lower performance for some subpopulations, particularly people with darker skin tones.  Wilson et al. investigated pedestrian detection with the BDD dataset and found poorer pedestrian prediction for darker skin tones that is not explained by confounding variables like time of day or occlusion \cite{wilson2019predictive}.  Wen et al. performed a systematic review of publicly available skin image datasets and found a substantial under-representation of darker skin types \cite{wen2021characteristics}. We believe that our approach could be used to recognize if images of people from underrepresented subpopulations map outside of the tested embedding region before harmful failures occur. Identifying, and potentially labeling, the operating images that are different than the images seen during training and testing is one way that our proposed approach could be used to improve DNN performance on underrepresented subpopulations.


\subsection{Limitations}\label{limitations}

We identify the tested regions in the embedding space with convex hulls. It is not clear that a convex hull is always an appropriate choice. It may be useful to measure the distance from the test samples to determine if samples are inside the tested region instead of a binary decision based on whether the operating samples are inside or outside the convex hull. Additionally, the convex hull may not be able to capture all kinds of corruptions or shifts of interest. For example,
we consider DNN classifiers that have been pre-trained with ImageNet; an adversarial sample that is a perturbed ImageNet image may still map within a convex hull if it maintains characteristics from the original dataset.

\section{Conclusions}
We propose a NGP method that can predict DNN performance in a novel operating domain without requiring labeled data, context distributions, or class distributions. We demonstrate the robustness of the method over three classification tasks, three DNN architectures, and five different realistic external datasets. In addition, we show that our proposed method to map the DNN embedding space can be leveraged for error prediction in tasks where the DNN performance is poor.
More broadly, we believe that NGP is a task that warrants more attention to enable
targeted performance prediction towards a specific operating domain. We believe this is a promising direction for further research and can be a step towards dependable and practical DNNs for safety critical tasks in unconstrained environments.

\subsection{Acknowledgements}
We would like to thank Jin Bai for his constructive input.
M.O. was supported through a Critical Path grant from the U.S. Food and Drug Administration, and by an appointment to the Research Participation Program at the Center for Devices and Radiological Health administered by the Oak Ridge Institute for Science and Education through an interagency agreement between the U.S. Department of Energy and the U.S. Food and Drug Administration. The contents of this work are solely the responsibility of the authors.


{\small
\bibliographystyle{ieee_fullname}
\bibliography{egbib}
}

\appendix

\end{document}